\pdfoutput=1

\documentclass[11pt]{article}

\usepackage{acl}

\usepackage{times}
\usepackage{latexsym}
\usepackage[T1]{fontenc}
\usepackage[utf8]{inputenc}

\usepackage{microtype}

\usepackage{url}
\usepackage{tipa}

\usepackage{graphicx}
\usepackage{amsmath}
\usepackage{amssymb}
\usepackage{amsfonts}
\usepackage{booktabs}
\usepackage{supertabular}
\usepackage{multirow}
\usepackage{lipsum}
\usepackage{enumitem}

\usepackage{caption}
\usepackage{subcaption}

\usepackage{color,soul}
\usepackage{xcolor}
\usepackage{linguex}

\usepackage{graphicx} 
\graphicspath{{./imgs/}}

\definecolor{mygreen}{RGB}{0, 150, 0}
\definecolor{myred}{RGB}{200, 0, 0}
\definecolor{myblue}{RGB}{0, 0, 200}
\definecolor{myorange}{RGB}{215, 107, 0}

\usepackage{cleveref}
\crefname{section}{\S}{\S\S}
\Crefname{section}{\S}{\S\S}
\crefname{table}{Tab.}{}
\crefname{figure}{Fig.}{Figs.}
\crefname{algorithm}{Algorithm}{}
\crefname{algorithm}{Algorithm}{}
\crefname{line}{Line}{}
\crefname{appendix}{App.}{}
\crefformat{section}{\S#2#1#3}
\crefname{thm}{Theorem}{}
\crefname{cor}{Corollary}{}
\crefname{prop}{Proposition}{}
\crefname{def}{Definition}{}

\newcommand{\citeposs}[1]{\citeauthor{#1}'s (\citeyear{#1})}
\newcommand{\defn}[1]{\textbf{#1}}

\DeclareMathOperator{\E}{\mathbb{E}}
\newcommand{\vinformation}{$\calV$-information}
\newcommand{\ent}{\mathrm{H}}
\newcommand{\MI}{\mathrm{I}}

\newcommand{\bWnull}{\mathbf{P}_\mathrm{null}}
\newcommand{\calV}{\mathcal{V}}
\newcommand{\vent}{\ent_{\calV}}
\newcommand{\vMI}{\MI_{\calV}}
\newcommand{\vUnc}{\mathrm{U}_{\calV}}
\newcommand{\R}{\mathbb{R}}
\newcommand{\bert}{\mathrm{model}}
\newcommand{\sentence}{\mathtt{sentence}}
\newcommand{\br}{\boldsymbol{r}}
\newcommand{\btheta}{\boldsymbol{\theta}}
\newcommand{\bthetainv}{\btheta^\intercal}
\newcommand{\ptheta}{p_{\btheta}}

\newcommand{\bA}{\mathbf{A}}
\newcommand{\bM}{\mathbf{M}}

\everypar={\looseness=-1}

\usepackage{inconsolata}

\usepackage{todonotes}

\title{Probing for the Usage of Grammatical Number}

\newcommand{\ucambridge}{\normalfont \text{\textipa{D}}}
\newcommand{\ethz}{\text{\normalfont \textipa{Q}}}
\newcommand{\lattice}{\normalfont \text{\textipa{@}}}
\newcommand{\unipi}{\normalfont \text{\textipa{B}}}

\author{Karim Lasri$^{\lattice,\unipi}$ Tiago Pimentel$^{\ucambridge}$ Alessandro Lenci$^{\unipi}$ Thierry Poibeau$^{\lattice}$ Ryan Cotterell$^{\ethz}$ \\
$^{\lattice}$Lattice (\'Ecole Normale Supérieure-PSL, CNRS, U. Sorbonne Nouvelle)~\; \\ $^{\unipi}$University of Pisa
~\;~$^{\ucambridge}$University of Cambridge
~\;~$^{\ethz}$ETH Z\"{u}rich \\
  \texttt{\href{mailto:karim.lasri@ens.psl.eu}{karim.lasri@ens.psl.eu}}~\;~ \texttt{\href{mailto:tp472@cam.ac.uk}{tp472@cam.ac.uk}} \\ \texttt{\href{mailto:alessandro.lenci@unipi.it}{alessandro.lenci@unipi.it}}~\;~ \texttt{\href{mailto:thierry.poibeau@ens.psl.eu}{thierry.poibeau@ens.psl.eu}} \\ \texttt{\href{mailto:ryan.cotterell@inf.ethz.ch}{ryan.cotterell@inf.ethz.ch}}
}

\date{}

\begin{document}

\maketitle

\begin{abstract}

A central quest of probing is to uncover \emph{how} pre-trained models encode a linguistic property within their representations.
An encoding, however, might be spurious---i.e., the model might not rely on it when making predictions.
In this paper, we try to find an encoding that the model actually uses, introducing a usage-based probing setup.
We first choose a behavioral task which cannot be solved without using the linguistic property.
Then, we attempt to remove the property by intervening on the model's representations.
We contend that, if an encoding is used by the model, its removal should harm the performance on the chosen behavioral task.
As a case study, we focus on how BERT encodes grammatical number, and on how it uses this encoding to solve the number agreement task.
Experimentally, we find that BERT relies on a linear encoding of grammatical number to produce the correct behavioral output.
We also find that BERT uses a separate encoding of grammatical number for nouns and verbs.
Finally, we identify in which layers information about grammatical number is transferred from a noun to its head verb.\looseness=-1
\end{abstract}

\section{Introduction}

Pre-trained language models have enabled researchers to build models that achieve impressive performance on a wide array of natural language processing (NLP) tasks \citep{devlin2018bert,liu,raffel2020exploring}.
How these models encode and use the linguistic information necessary to perform these tasks, however, remains a mystery.
Over recent years, a number of works have tried to demystify the inner workings of various pre-trained language models \citep{alain2016understanding,adi2017finegrained,elazar2021amnesic}, but no comprehensive understanding of how the models work has emerged.
Such analysis methods are typically termed \emph{probing}, and are methodologically diverse.\looseness=-1

In our assessment, most research in probing can be taxonomized into three distinct paradigms.
In the first paradigm, \defn{diagnostic probing}, researchers typically train a supervised classifier to predict a linguistic property from the models' representations. 
High accuracy is then interpreted as an indication that the representations encode information about the property \citep{alain2016understanding, adi2017finegrained, hupkes, conneau-etal-2018-cram}.
A second family of methods, \defn{behavioral probing},
consists in observing a model's behavior directly, typically studying the model's predictions on hand-picked evaluation datasets \citep{linzen-etal-2016-assessing,goldberg,warstadt-etal-2020-learning,ettinger-2020-bert}.
Finally, \defn{causal probing} methods rely on
interventions to evaluate how specific components impact a model's predictions \citep{giulianelli-etal-2018-hood, vig, elazar2021amnesic}.%

In this paper, we will investigate how linguistic properties are encoded in a model's representations, where we use the term encoding to mean the subspace on which a model relies to extract---or decode---the information.
While probing has been extensively used to investigate whether a linguistic property is encoded in a set of representations,
it still cannot definitively answer whether a model actually uses a certain encoding.
Diagnostic probes, for instance, may pick up on a \defn{spurious encoding} of a linguistic property, i.e., an encoding that allows us to extract our target property from the representation, but which the model being probed may not actually use to make a prediction.

Combining the three paradigms above, we instead seek to find encodings that are actually used by a pre-trained model, which we term \defn{functional encodings}.
To that end, we take a usage-based perspective on probing.
Under this perspective, a researcher first identifies a linguistic property to investigate (e.g.,  grammatical number), and selects a behavioral task which requires knowledge of this property (e.g., selecting a verb's inflection which agrees in number with its subject).
The researcher then performs a causal intervention with the goal of removing a specific encoding (of the linguistic property under consideration) from the model's representations.
If the encoding is a functional encoding, i.e., an encoding that the model indeed uses to make a prediction, then the intervention should prevent the model from solving the task.\footnote{
See \citet{ravfogel-etal-2021-counterfactual} for a similar pipeline.
}
Finally, once a functional encoding is discovered, we can use it to track how the property's information flows through the model under investigation.

\vspace{-2pt}
As a case study, we examine how BERT \cite{devlin2018bert} uses grammatical number to solve a number agreement task.
In English, grammatical number is a binary morpho-syntactic property: A word is plural or singular.
In turn, subject--verb number agreement is a behavioral task; it inspects whether a model can predict the correct verbal inflection given its subject's number.
For a model to solve the task, it thus requires information about the grammatical number of the subject and the verb.
Our goal is to find how the model encodes this information when using it to make predictions. 
In other words, we want to find the structure from which the model \emph{decodes} number information when solving the task.\looseness=-1

\vspace{-2pt}
In our experiments, we make three findings.
First, our experiments provide us with strong evidence that BERT relies on a linear functional encoding of grammatical number to solve the number agreement task.
Second, we find that nouns and verbs do \emph{not} have a shared functional encoding of number; in fact, BERT relies on disjoint sub-spaces to extract their information.
Third, our usage-based perspective allows us to identify where number information (again, as used by our model to make predictions) is transferred from a noun to its head verb.
Specifically, we find that this transfer occurs between BERT's $3^{\text{rd}}$ and $8^{\text{th}}$ layers, and that most of this information is passed indirectly through other tokens in the sentence.

\vspace{5pt}
\section{Paradigms in Probing}\label{sec:probing_methodologies}
A variety of approaches to probing have been proposed in the literature.
In this paper, we taxonomize them into three paradigms:
(i) diagnostic probing, (ii) behavioral probing, and (iii) causal probing.

\paragraph{Diagnostic Probing.} 
Traditionally, probing papers focus on training supervised models on top of frozen pre-trained representations \citep{adi2017finegrained,maudslay-etal-2020-tale}.
The general assumption behind the work is that, if a probe achieves high accuracy, then the property of interest is encoded in the representations.
Many researchers have expressed a preference for linear classifiers in probing \cite{alain2016understanding,ettinger-etal-2016-probing,hewitt-manning-2019-structural}, suggesting that a less complex classifier gives us more insight into the model.
Others, however, called this criterion into question \citep{tenney-etal-2019-bert,tenney2018what,voita-titov-2020-information,papadimitriou-etal-2021-deep,sinha2021masked,pimentel-etal-2020-pareto,pimentel-cotterell-2021-bayesian}.
Notably, \citet{hewitt-liang-2019-designing} proposed that complex classifiers may learn to extract a property by themselves, and may thus not reflect any true pattern in the representations.
Further, \citet{pimentel-etal-2020-information} showed that, under a weak assumption, contextual representations encode as much information as the original sentences.
Ergo, it is not clear what we can conclude from diagnostic probing alone.

\paragraph{Behavioral Probing.} 
Another probing paradigm analyzes the behavior of pre-trained models on carefully curated datasets.
The researcher analyzes the model's predictions on them.
By avoiding the use of diagnostic probes, they do not fall prey to the criticism above---tasks are directly performed by the model, and thus must reflect the pre-trained models' acuity.
One notable example is \citet{linzen-etal-2016-assessing}, who evaluate a language model's syntactic ability via a careful analysis of a number agreement task. 
By controlling the evaluation, \citeauthor{linzen-etal-2016-assessing} could disentangle the model's syntactic knowledge from a heuristic based on linear ordering.
In a similar vein, a host of recent work makes use of carefully designed test sets to perform behavioral analysis \citep{ ribeiro-etal-2020-beyond,warstadt-etal-2020-learning,warstadt2020can,lovering2021predicting, newman2021refining}.
While behavioral probing often yields useful insights, the paradigm typically treats the model itself as a blackbox, thus failing to explain how individual components of the model work.\looseness=-1

\paragraph{Causal Probing.}
Finally, the third probing paradigm relies on causal interventions \citep{vig,tucker-etal-2021-modified,ravfogel-etal-2021-counterfactual}.
In short, the researcher performs causal interventions that modify parts of the network during a forward pass (e.g., a layer's hidden representations) to determine their function.
For example, \citet{vig2020investigating} fix a neuron's value while manipulating the model's input to evaluate this neuron's role in mediating gender bias.
Relatedly, \citet{elazar2021amnesic} propose a method to erase a target property from a model's intermediate layers.
They then analyze the effect of such interventions on a masked language model's outputs.\looseness=-1

\section{Probing for Usage}
Under our usage-based perspective, our goal is to find a functional encoding---i.e., an encoding that the model actually uses when making predictions.
We achieve this by relying on a combination of the paradigms discussed in \cref{sec:probing_methodologies}.
To this end, we first need a behavioral task that requires the model to use information about the target property.
We then perform a causal intervention to try to remove this property's encoding.
We explain both these components in more detail now.

\paragraph{Behavioral Task.}
We first require a behavioral task which can \emph{only} be solved with information about the target property.
The choice of task and target property are thus co-dependent.
Further, we require our model to perform well on this task.
Why? If the model cannot achieve high performance on the behavioral task, we cannot be sure the model encodes the target property, e.g., grammatical number, at all. 
However, if the model can perform the task, we contend it must make use of the property.\looseness=-1
\paragraph{Causal Intervention.} 
Our goal in this work is to answer a causal question: Can we identify a property's functional encoding?
We thus require a way to intervene in the model's representations.
If a model relies on an encoding to make predictions, removing it should harm the model's performance on the behavioral task.
It follows that, by measuring the impact of our interventions on the model's behavioral output, we can assess whether our model was indeed decoding information from our targeted encoding.

\section{Grammatical Number and its Usage}

The empirical portion of this paper focuses on a study of how BERT encodes grammatical number in English.
We choose number as our object of study because it is a well understood morpho-syntactic property in English.
Thus, we are able to formulate simple hypotheses about how BERT passes information about number when performing number agreement.
We use \citeposs{linzen-etal-2016-assessing} number agreement task as our behavioral task.\looseness=-1

\subsection{The Number Agreement Task} 
\label{sec:number-agreement}

In English, a verb and its subject agree in grammatical number \citep{CorbettGG2006A}.
Consider, for instance, the sentences: 

\begin{small}

\ex.\label{ex1}
\a.\label{ex1a} \phantom{*}The  \textcolor{blue}{boy} \textcolor{orange}{goes} to the movies.
\b. \label{ex1b} *The  \textcolor{blue}{boy} \textcolor{orange}{go} to the movies.
\c. \label{ex1c} \phantom{*}The  \textcolor{blue}{boy} that holds the \emph{keys} \textcolor{orange}{goes} to the movies.
\d. \label{ex1d} *The \textcolor{blue}{boy} that holds the \emph{keys}  \textcolor{orange}{go} to the movies.

\end{small}

\noindent In the above sentences, both \ref{ex1a} and \ref{ex1c} are grammatical, but \ref{ex1b} and \ref{ex1d} are not; this is because, in the latter two sentences, the highlighted verb does not agree in number with its subject.

The subject--verb number agreement task evaluates a model's ability to predict the correct verbal inflection, measuring its preference for the grammatical sentence.
In this task, the probed model is typically asked to predict the verb's number given its context.
The model is then considered successful if it assigns a larger probability to the correct verb inflection:\looseness=-1
\begin{itemize}[leftmargin=2pt,itemsep=1pt]
\begin{small}
 \item[] \textbf{context}: The \textcolor{blue}{boy} that holds the keys \textcolor{orange}{[MASK]} to the movies.
 \item[] \textbf{success}: $p(\textcolor{orange}{goes} \mid \textit{context}) > p(\textcolor{orange}{go}\mid \textit{context})$
 \item[] \textbf{failure}: $p(\textcolor{orange}{go} \mid \textit{context}) > p(\textcolor{orange}{goes}\mid \textit{context})$
\end{small}
\end{itemize}
In this setting, the subject is usually called the \textbf{cue} of the agreement, and the verb is called the \textbf{target}.\looseness=-1

Examples similar to the above are often designed to study the impact of distractors (the word \emph{keys} in \ref{ex1c} and \ref{ex1d}) on the model's ability to predict the correct verb form.
Success on the task is usually taken as evidence that a model is able to track syntactic dependencies. In this regard, this phenomenon has been studied in a variety of settings to investigate the syntactic abilities of neural language models \citep{gulordava-etal-2018-colorless, marvin-linzen-2018-targeted, newman2021refining, lasri}.
In this work, however, we do not use this task to make claims about the syntactic abilities of the model, as done by \citet{linzen-etal-2016-assessing}.
Instead, we employ it as a case study to investigate how BERT encodes and uses grammatical number.\looseness=-1

\subsection{Related Work on Grammatical Number}

A number of studies have investigated how grammatical number is encoded in neural language models.\footnote{We focus on grammatical number here. There is, however, also a vast literature investigating how BERT encodes number from a numeracy point of view \citep{wallace-etal-2019-nlp, geva-etal-2020-injecting, spithourakis-riedel-2018-numeracy}.} 
Most of this work, however, focuses on diagnostic probes \citep{klafka-ettinger-2020-spying,torroba-hennigen-etal-2020-intrinsic}.
These studies are thus agnostic about whether the probed models actually use the encodings of number they discover.
Some authors, however, do consider the relationship between how the model encodes grammatical number and its predictions.
Notably, \citet{giulianelli-etal-2018-hood} use a diagnostic probe to investigate how an LSTM encodes number in a subject--verb number agreement setting.
Other approaches \citep{lakretz-etal-2019-emergence, finlayson-etal-2021-causal} have been proposed to apply interventions at the neuron level and track their effect on number agreement.
In this work, we look for functional encodings of grammatical number---encodings which are in fact used by our probed model when solving the task.

\section{From Encoding to Usage}

We discuss how to identify and remove an encoding from a set of contextual representations using diagnostic probing.
Our use of diagnostic probing is thus twofold.
For a model to rely on an encoding of our property when making predictions, the property must be encoded in its representations.
We thus first use diagnostic probing to measure the amount of information a representation contains about the target linguistic property.
In this sense, diagnostic probing serves to sanity-check our experiments---if we cannot extract information from the representations, there is no point in going forward with our analysis.
Second, we make use of diagnostic probing in the context of amnesic probing \citep{elazar2021amnesic}, which allows us to determine whether this probe finds a functional or a spurious encoding of the target property.

\subsection{Estimating Extractable Information} \label{sec:diagnostic_classifiers}

In this section, we discuss how to estimate the amount of extractable number information in our probed model's  representations.
This is the probing perspective taken by \citet{pimentel-etal-2020-information} and \citet{hewitt-etal-2021-conditional} in their diagnostic probing analyses.
The crux of our analysis relies on the fact that the encoding extracted by diagnostic probes is not necessarily the functional encoding used by our probed model.
Nevertheless, for a model to use a property in its predictions, this property should at least be extractable, which is true due to the data processing inequality.
In other words, extractability is a necessary, but not sufficient, condition for a property to be used by the model.
We quantify the amount of extractable information in a set of representations in terms of a diagnostic probe's \vinformation{} \citep{xu2020theory}, where the \vinformation{} is a direct measure of the amount of extractable information in a random variable. 
We compute the \vinformation{} as:\footnote{
See \cref{app:v-information} for a detailed description of \vinformation{}. 
}%
\begin{equation}
    \vMI(R \rightarrow N) = \vent(N) - \vent(N \mid R)
\end{equation}
where $R$ and $N$ are, respectively, a representation-valued and a number-valued random variables, $\mathcal{V}$ is a variational family determined by our diagnostic probe,
and the $\calV$-entropies are defined as:%
\begin{align}
    \vent(N) &= \inf_{q \in \calV} \E_{n \sim p(n)} \log \frac{1}{q(n)} \\
    \vent(N \mid R) &= \inf_{q \in \calV} \E\limits_{n, \br \sim p(n, \br)} \log \frac{1}{q(n \mid \br)}
\end{align}
where $p$ is the true joint distribution over number and representations.\footnote{Representations are model-specific. 
However, the true distributions over representations here refers to the distribution over representations that is induced by the true distribution over natural language sentences.}

Further, if we denote our analyzed model's (i.e., BERT's) hidden representations as:\looseness=-1
\begin{equation}
    \br_{t, l} = \bert(\sentence)_{t, l}
\end{equation}
we define our linear diagnostic probe as:
\begin{equation} \label{eq:linear_pred}
    \ptheta(N = \textsc{sg} \mid R = \br_{t, l}) = \sigma(\bthetainv\, \br_{t,l} + b)
\end{equation}
where $\br_{t, l} \in \R^{768}$ is a column vector, $t$ is a sentence position and $l$ is a layer,
$N$ is the binary number label associated with the word at position $t$, $\sigma$ is the sigmoid function, $\btheta$ is a real-valued column parameter vector and $b$ is a bias term.
In this case, we can define our variational family as $\calV = \{\ptheta \mid \btheta \in \R^{768}\}$.

\subsection{Intervening on the Representations}
\label{sec:amnesic-probing}

We now discuss how we perform a causal intervention to prevent the analyzed model from using a given encoding.
The goal is to damage the model and make it ``forget'' a property's information.
This allows us to analyze whether that encoding actually influences the probed model's predictions---i.e., whether this encoding is indeed functional.
To this end, we employ amnesic probing \citep{elazar2021amnesic}.\footnote{In particular, this intervention consists in applying iterative null-space projection to the representations, originally proposed by \citet{ravfogel-etal-2020-null}. We note that \citet{ravfogel2022linear,ravfogel2022adversarial} recently proposed two new methods to remove information from a set of representations.}
In short, we first learn a linear diagnostic classifier, following \cref{eq:linear_pred}. 
We then compute the projector onto the kernel (or null) space of this linear transform $\btheta$, shown below:
\begin{equation}
    \bWnull = \mathbf{I} - \frac{\btheta\btheta^{\intercal}}{||\btheta||_2^2} 
\end{equation}
By iterating this process, we store a set of parameter vectors $\btheta^{(k)}$ and their associated projectors $\bWnull^{(k)}$ until we are unable to extract the property.
The composition of these projectors makes it possible to remove all linearly extractable number information from the analyzed representations. 
We can then apply the resulting composition to the said representations to get a new set of vectors:
\begin{equation}
    \br_{t,l}' = \bWnull^{(k)}\, \cdots\, \bWnull^{(2)}\, \bWnull^{(1)}\, \br_{t,l}
\end{equation}
After learning the projectors, we can measure how erasing a layer's encoding impacts: (i) the subsequent layers, and (ii) our model's performance on the number agreement task.
Removing a functional encoding of grammatical number should cause a performance drop on the number agreement task. 
Further, looking at both (i) and (ii) allows us to make a connection between the amount of information we can extract from our probed model's layers and its behavior.
We are thus able to determine whether the encodings revealed by our diagnostic probes are valid from a usage-based perspective---are they actually used by the probed model on a task that requires them?%

\section{Experimental Setup}

\paragraph{Data.}
We perform our analysis on \citeposs{linzen-etal-2016-assessing} number agreement dataset, which consists of sentences extracted from Wikipedia.
In this dataset, each sentence has been labeled with the position of the cue and target, along with their grammatical number.
We assume here that this dataset is representative of the number agreement task; this may not be true in general, however.

\paragraph{Model.}%
In our experiments, we probe BERT \citep{devlin2018bert}.\footnote{We focus on \texttt{bert-base-uncased}, as implemented in the transformers library \citep{wolf-etal-2020-transformers}.}
Specifically, BERT is a bidirectional transformer model with 12 layers, trained using a masked language modeling objective.
As BERT has been shown to perform well on this dataset \citep{goldberg}, we already know that our probed model passes our first requirement; BERT does use number information in its predictions.

\paragraph{Distinguishing Nouns and Verbs.}
\label{sec:probed_representations}
While number is a morpho-syntactic property common to nouns and verbs, we do not know \textit{a priori} if BERT relies on a single subspace to encode number in their representations.
Though it is possible for BERT to use the same encoding, it is equally plausible that each part of speech would get its own number encoding.
This leads us to perform our analyses using independent sets of representations for nouns and verbs; as well as a mixed set which merges both of them. 
Further, verbs are masked when performing the number agreement task, so their representations differ from those of unmasked verbs. 
Ergo, we analyze both unmasked and masked tokens at the target verb's position---which for simplicity we call verbs and masked verbs, respectively.
This leaves us with four probed categories: 
\textcolor{myblue}{nouns}, \textcolor{myorange}{verbs}, \textcolor{mygreen}{masked-verbs}, and \textcolor{myred}{mixed}.\looseness=-1

\begin{figure*}
    \centering
     \begin{subfigure}[b]{0.32\textwidth}
         \centering
         \includegraphics[width=\textwidth]{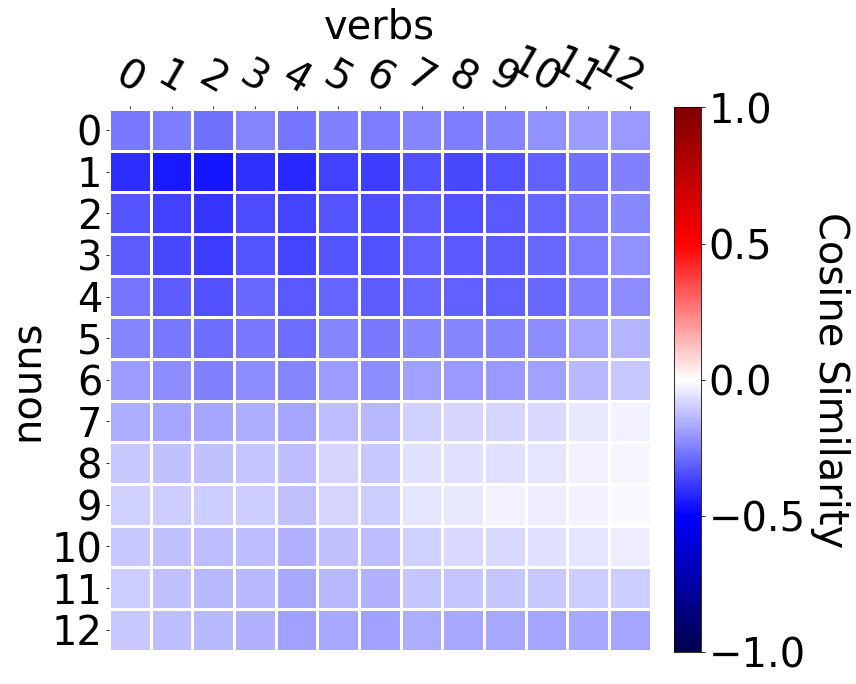}
     \end{subfigure}%
     ~\hfill
     \begin{subfigure}[b]{0.32\textwidth}
         \centering
         \includegraphics[width=\textwidth]{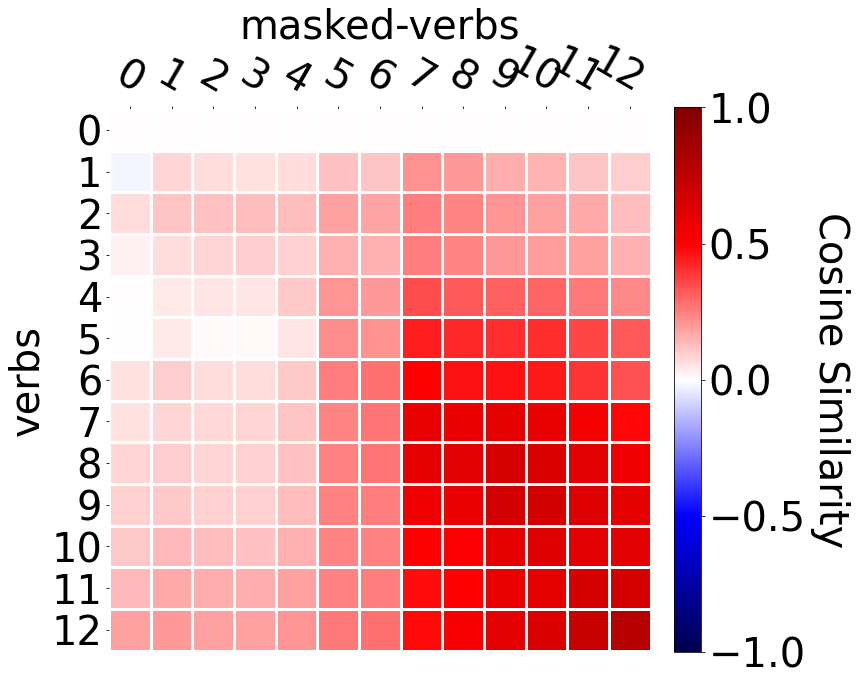}
     \end{subfigure}%
     ~\hfill
     \begin{subfigure}[b]{0.32\textwidth}
         \centering
         \includegraphics[width=\textwidth]{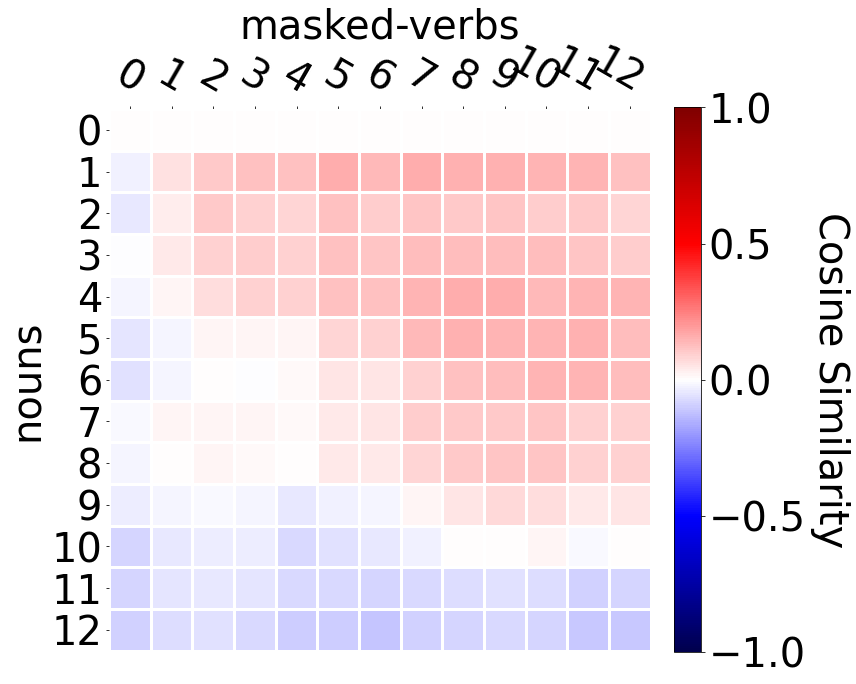}
     \end{subfigure}
    \caption{Cosine similarities between the learned parameter vectors of our diagnostic probes. The matrices display similarities between different layers, and across categories.}
    \label{fig:cosine-distance-matrices}
    \vspace{-5pt}
\end{figure*}

\section{Experiments and Results}
In our experiments, we focus on  %
answering two questions: (i) How is number information encoded in BERT's representations? and (ii) How is number information transferred from a noun to its head verb for the model to use it on the behavioral task?
We answer question (i) under both extractability
and usage-based perspectives.
In \cref{sec:results_diagnostic}, we present our sanity-check experiments that demonstrate that grammatical number is indeed linearly extractable from BERT's representations.
In \cref{sec:used_encodings} and \cref{sec:shared_encodings}, we use our causal interventions: we identify BERT's \emph{functional encodings} of number; and analyze whether these functional encodings are shared across parts of speech.
Finally, in \cref{sec:amnesic_transfer} and \cref{sec:attention_transfer} we investigate question (ii), taking a closer look at the layers in which information is passed.

\begin{figure}
    \centering
    \includegraphics[width=\columnwidth]{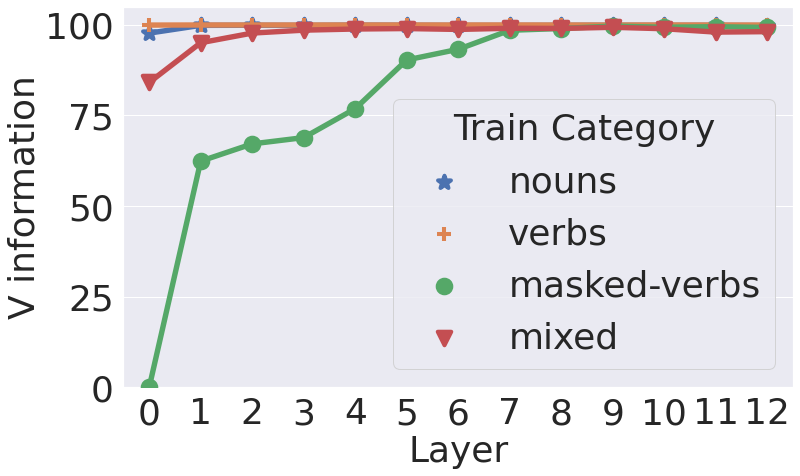}
    \caption{The amount of \vinformation{} BERT representations hold about grammatical number, as estimated with linear diagnostic probes.\looseness=-1}
    \label{fig:sanity_check}
\end{figure}

\subsection{What do diagnostic probes say about number?} \label{sec:results_diagnostic}

\cref{fig:sanity_check} presents diagnostic probing results in all four of our analyzed settings.\footnote{
We further present accuracy results in \cref{sec:extra_results}.}
\textit{A priori}, we expect that verbs' and nouns' representations should already contain a large amount of \vinformation{} about their grammatical number at the type level.
As expected, we see that the $\calV$-information is near its maximum for both verbs and nouns in all layers; this means that nearly 100\% of the uncertainty about grammatical number is eliminated given BERT's representations.
Further, the mixed category results also reach a maximal \vinformation{}, which indicates that it is possible to extract information linearly about both categories at the same time.
On the other hand, the \vinformation{} of masked verbs is 0 at the non-contextual layer and it progressively grows as we get to the upper layers.\footnote{
We note that, in \cref{fig:sanity_check}, layer 0 corresponds to the non-contextual representations (i.e. the word embeddings before being summed to BERT's position embeddings).
Non-contextual layers thus contain no information about the number of a masked verb, as the mask token contains no information about its replaced verb's number.}
As we go to BERT's deeper layers, the \vinformation{} steadily rises, with nearly all of the original uncertainty eliminated  in the mid layers. 
This suggests that masked verbs' representations acquire number information in the first 7 layers.\looseness=-1

However, from these results alone we cannot confirm whether the encoding that nouns and verbs use for number is shared or disjoint.
We thus inspect the encoding found by our diagnostic probes, evaluating the cosine similarity between their learned parameters $\btheta$ (ignoring the probes' bias terms $b$ here).
If there is a single shared encoding across categories, these cosine similarities should be high. 
If not, they should be roughly 0.
\cref{fig:cosine-distance-matrices} (left) shows that nouns and verbs might encode number along different directions.
Specifically, noun representations on the first 6 layers seem to have a rather opposite encoding from verbs, while the later layers are mostly orthogonal. 
Further, while masked verbs and verbs do not seem to share an encoding in the first few layers, they are strongly aligned from layer 6 on (\cref{fig:cosine-distance-matrices}; center).

We now know that there are encodings from which we can extract number from nouns and verbs, and that these encodings are disjoint. However, we still do not know whether the encoding is spurious or functional.

\begin{figure*}
    \begin{subfigure}[b]{0.48\textwidth}
         \centering
         \includegraphics[width=\columnwidth]{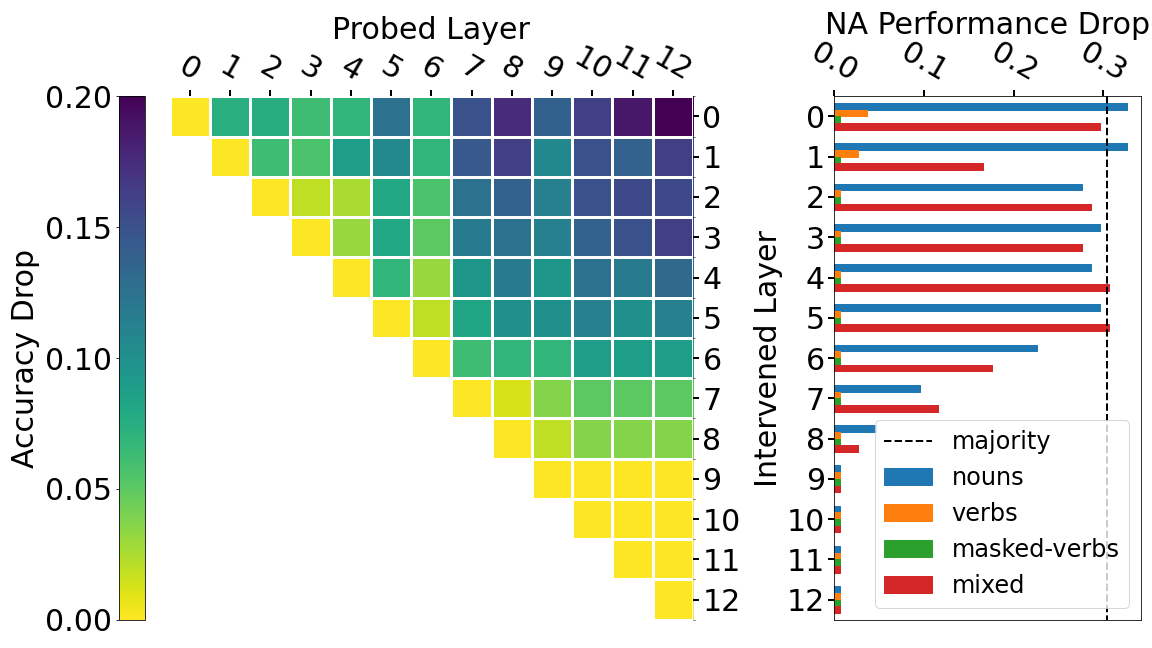}
         \subfloat[\label{fig:amnesic-retrieval-cue} Information loss (measured at the target) after erasing nouns' number information at the cue position.\looseness=-1]{\hspace{.6\linewidth}}
         \hfill
         \subfloat[\label{fig:amnesic-agreement-cue} NA performance drop after erasing number at the cue position.]{\hspace{.33\linewidth}}
     \end{subfigure}
     ~
    \hfill
    \centering
     \begin{subfigure}[b]{0.48\textwidth}
        \centering
        \includegraphics[width=\columnwidth]{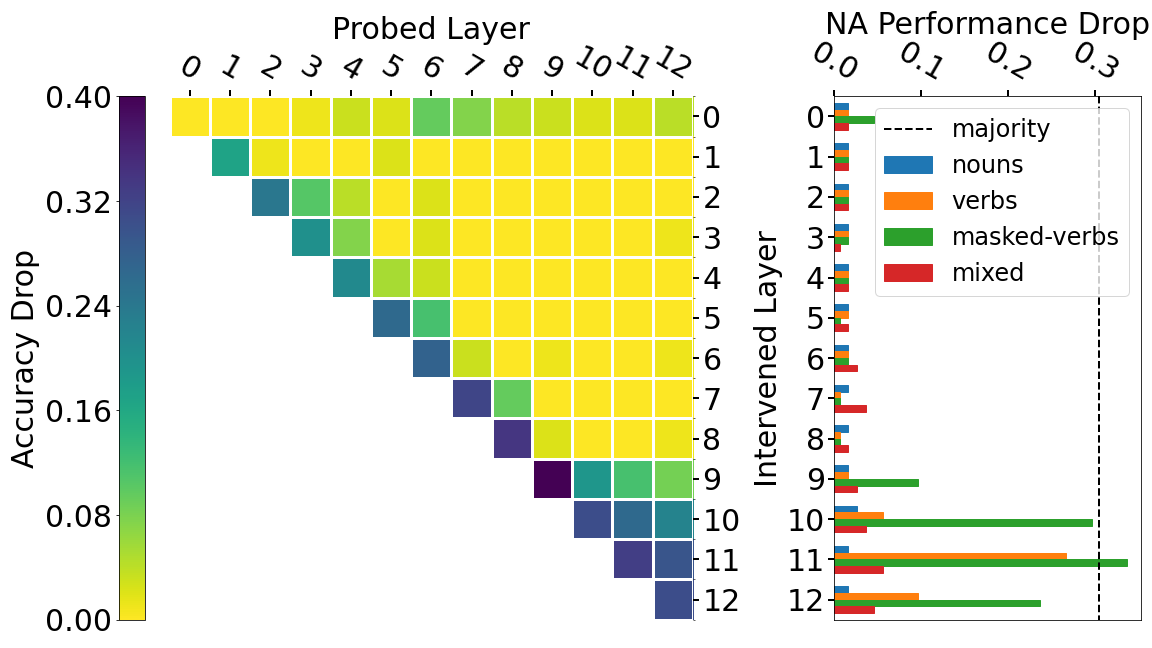}
        \subfloat[\label{fig:amnesic-retrieval-target} Information loss (measured at the target) after erasing masked verbs' number at the target position.]{\hspace{.6\linewidth}}
        \hfill
        \subfloat[\label{fig:amnesic-agreement-target} NA performance drop after erasing number at the target position.]{\hspace{.33\linewidth}}
     \end{subfigure}
    \caption{Effect of our causal interventions on information recovery in subsequent layers (triangular matrices) and on the number agreement task (bar charts). Information loss is measured at the target position by a diagnostic probe; we display the probing accuracy drop compared to when no intervention was performed.
    The legend in the bar charts indicates what category the amnesic projectors have been trained on.
    Majority represents the difference in performance between BERT and a trivial baseline which always guesses the majority label.
    }
    \label{fig:amnesic-probe-recovery-matrices}
\end{figure*}

\subsection{Does the model use these encodings?} \label{sec:used_encodings}

The patterns previously observed suggest there is a linear encoding, from which grammatical number can be extracted from BERT's representations.
We, however, cannot determine whether these encodings are actually those used by the model to make predictions.
We now answer this question taking our proposed usage-based perspective, studying the impact of linearly removing number information at both the cue and target positions.\footnote{The number of dimensions removed by our amnesic projectors in each layer and category is presented in \cref{tab:amnesic-baselines}.\looseness=-1}
We evaluate the model's change in behavior, as evaluated by its performance on the number agreement (NA) task.

\cref{fig:amnesic-retrieval-cue} and \cref{fig:amnesic-retrieval-target} show the decrease in how much information is extractable at the target position after the interventions are applied. 
\cref{fig:amnesic-agreement-cue} and \cref{fig:amnesic-agreement-target} show BERT's accuracy drops on the NA task (as measured at the output level).
By comparing these results, we find a strong alignment between the information lost across layers and the damage caused to the performance on the task---irreversible information losses resulting from our intervention are mirrored by a performance decrease on the NA task.
This alignment confirms that the model indeed uses the linear information erased by our probes.
In other words, we have found the probed property's functional encoding.\looseness=-1

\subsection{Does BERT use the same encoding for verbs and nouns?} \label{sec:shared_encodings}

We now return to the question of whether nouns and verbs share a functional encoding of number, or whether BERT encodes number differently for them.
To answer this question, we investigate the impact of removing a category's encoding from another category, e.g., applying an amnesic projector learned on verbs to a noun.
In particular, we measure how these interventions decrease BERT's performance in our behavioral task.
\cref{fig:amnesic-agreement-cue,fig:amnesic-agreement-target} presents these results.

We observe that each category's projector has a different impact on performance depending on whether it is applied to the cue or the target.
\cref{fig:amnesic-agreement-cue}, for instance, shows that using the verb's, or masked verb's, projector to erase information at the cue's (i.e., the noun's) position does not hurt the model.
It is similarly unimpactful (as shown in \cref{fig:amnesic-agreement-target}) to use the noun's projectors to erase a target's (i.e., the masked verb's) number information. 
Further, the projector learned on the mixed set of representations does affect the cue, but has little effect on the target.
Together, these results confirm that BERT relies on rather distinct encodings of number information for nouns and verbs.\footnote{
A potential criticism of amnesic probing is that it may remove more information than necessary.
Cross-testing our amnesic probes, however, results in little effect on BERT's behavior.
It is thus likely that they are not overly harming our model.
Further, we also run a control experiment proposed by \citeauthor{elazar2021amnesic}, removing random directions at each layer (instead of the ones found by our amnesic probes). These results are displayed in the appendix in \cref{tab:amnesic-baselines}.}\looseness=-1

These experiments allow us to make stronger claims about BERT's encoding of number information.
First, the fact that our interventions have a direct impact on BERT's behavioral output confirms that the encoding we erase actually bears number information \emph{as used by the model when making predictions}.
Second, the observation from \cref{fig:cosine-distance-matrices}---that number information could be encoded orthogonally for nouns and verbs---is confirmed from a usage-based perspective.
Indeed, using amnesic probes trained on nouns has no impact when applied to masked verbs, and amnesic probes trained on verbs have no impact when applied to nouns.
These fine-grained differences in encoding may affect larger-scale probing studies if one's goal is to understand the inner functioning of a model.
Together, these results invite us to employ diagnostic probes more carefully, as the encoding found may not be actually used by the model.

\subsection{Where does number erasure affect the model?} \label{sec:amnesic_transfer}

Once we have found which encoding the model uses, we can pinpoint at which layers the information is passed from the cue to the target. 
To that end, we observe how interventions applied in each layer affect performance.
We know number information must be passed from the cue to the target's representations---otherwise the model cannot solve the task. 
Therefore, applying causal interventions to remove number information should harm the model's behavioral performance when applied to: (i) the cue's representations before the transfer occurs; (ii) the target's representations after the transfer occurred.

Interestingly, we observe that target interventions are only harmful after the 9$^{\mathrm{th}}$ layer; while noun interventions only hurt up to the 8$^{\mathrm{th}}$ layer (again, shown in \cref{fig:amnesic-probe-recovery-matrices}).
This suggests that the cue passes its number information in the first 8 layers, and that the target stops acquiring number information in the last three layers. 
While we see a clear stop in the transfer of information after layer 8, \cref{fig:amnesic-retrieval-cue} shows that the previous layers' contribution decreases slowly up to that layer. 
We thus conclude that information is passed in the layers before layer 8; however, we concede that our analysis alone makes it difficult to pinpoint exactly which layers.

\subsection{Where does attention pruning affect number transfer?} \label{sec:attention_transfer}

\begin{figure}
    \centering
     \begin{subfigure}[b]{0.48\columnwidth}
         \centering
         \includegraphics[width=\textwidth]{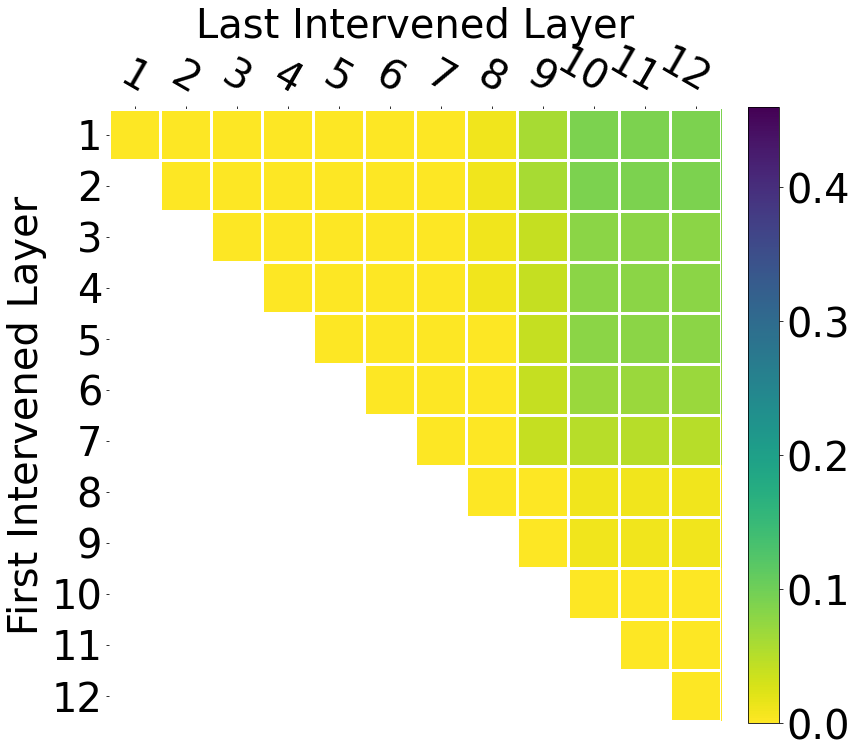}
         \caption{Removing attention from the target to the cue only}
     \end{subfigure}
     ~
     \begin{subfigure}[b]{0.48\columnwidth}
         \centering
         \includegraphics[width=\textwidth]{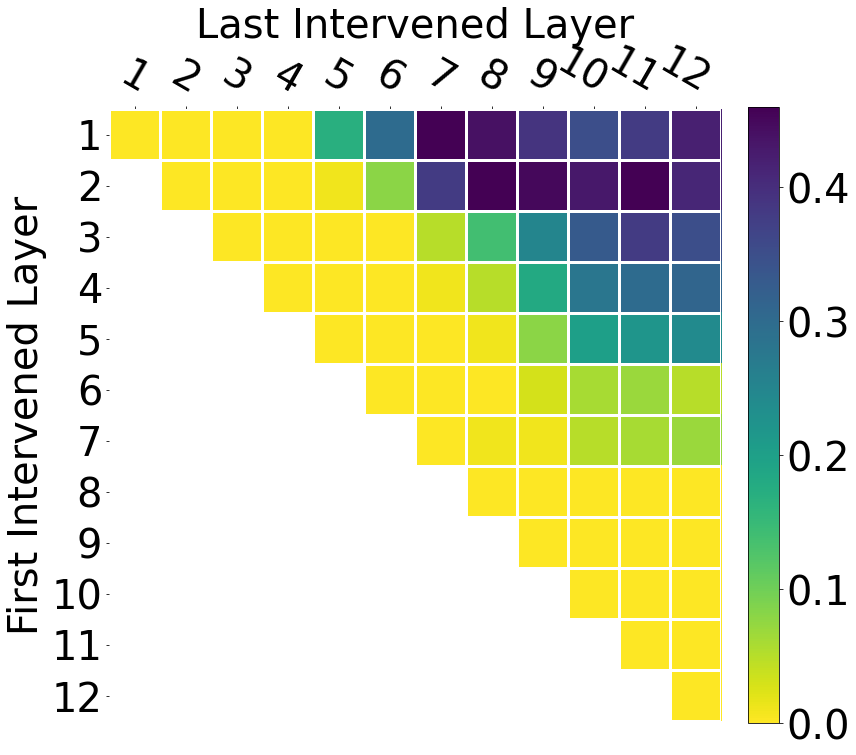}
         \caption{Removing attention from all tokens to the cue}
     \end{subfigure}
     \vspace{3pt}
    \caption{Number agreement task performance drops after performing attention removal. The attention cut is performed on a range of layers. Rows and columns, respectively, represent the first and last intervened layer.}
    \label{fig:attention-intervention-range-matrices}
\end{figure}

Finally, in our last experiments, we complement our analysis by performing attention removal to investigate how and where information is transmitted from the cue to the target position.
This causal intervention first serves the purpose of identifying the layers where information is transmitted.
Further, we wish to understand whether information is passed directly, or through intermediary tokens.
To this end, we look at the effect on NA performance after: (i) cutting direct attention from the target to the cue at specific layers, (ii) cutting attention from all tokens to the cue (as information could be first passed to intermediate tokens, which the target could attend to in subsequent layers).\footnote{\citet{klafka-ettinger-2020-spying}, for instance, showed that number information of a given token was distributed to neighboring tokens in the upper layers.}
Specifically, we perform these interventions in ranges of layers (from layer $i$ up to $j$). 
We report number agreement accuracy drops in \cref{fig:attention-intervention-range-matrices}.\footnote{We detail these interventions in \cref{sec:attention_intervention}.}

The diagonals from this figure show that removing attention from a single layer has basically no effect. 
Further, cutting attention from layers 6 to 10 suffices to observe near-maximal effect for direct attention.
Interestingly, it is at those layers where we see a transition from it being more harmful to apply amnesic projectors to the cue or to the target (in \cref{sec:amnesic_transfer}).
However, while those layers play a role in carrying number information to the target position, the drop is relatively modest when cutting only direct attention ($\approx 10\%$).
Cutting attention from all tokens to the cue, in turn, has a significant effect on performance (up to $\approx 40\%$), and is maximal for layers from 2 to 8. 
This first suggests that, while other clues in the sentence could indicate the target verb's number (such as a noun's determiner), the noun itself is the core source of number information.
Further, this shows the target can get information from intermediate tokens, instead of number being passed exclusively through direct attention.\footnote{See \cref{app:linear_study} for further experiments.}\looseness=-1

\section{Conclusion}
Our analysis of grammatical number allows us to track how a simple morpho-syntactic property, grammatical number, is encoded across BERT's layers and where it is transferred between them before being used on the model's predictions.
Using carefully chosen causal interventions, we demonstrate that forgetting number information impacts both: (i) BERT's behavior and (ii) how much information is extractable from BERT's inner layers.
Further, the effects of our interventions on these two, i.e., behavior and information extractability, line up satisfyingly, and reveal the encoding of number to be orthogonal for nouns and verbs.
Finally, we are also able to identify the layers in which the transfer of information occurs, and find that the information is not passed directly but through intermediate tokens.
Our ability to concretely evaluate our interventions' impact is due to our focus on grammatical number and the number agreement task which directly aligns probed information and behavioral performance.

\section*{Ethics Statement}
The authors foresee no ethical concerns with the work presented in this paper.

\section*{Acknowledgments}

We thank Josef Valvoda, the anonymous reviewers, and the meta-reviewer, for their invaluable feedback in improving this paper.
Karim Lasri's work is funded by the French government under management of Agence Nationale de la Recherche as part of the ``Investissements d'avenir'' program, reference ANR-19-P3IA-0001 (PRAIRIE 3IA Institute).
Ryan Cotterell acknowledges support from the Swiss National Science Foundation (SNSF) as part of the ``The Forgotten Role of Inductive Bias in Interpretability'' project.
Tiago Pimentel is supported by a Meta Research PhD Fellowship.

\bibliographystyle{acl_natbib}
\bibliography{bibliography}

\appendix

\onecolumn 

\section{Diagnostic Probing Cross-Evaluation}\label{sec:extra_results}

\begin{figure}[h]
        \centering
        \includegraphics[width=\textwidth]{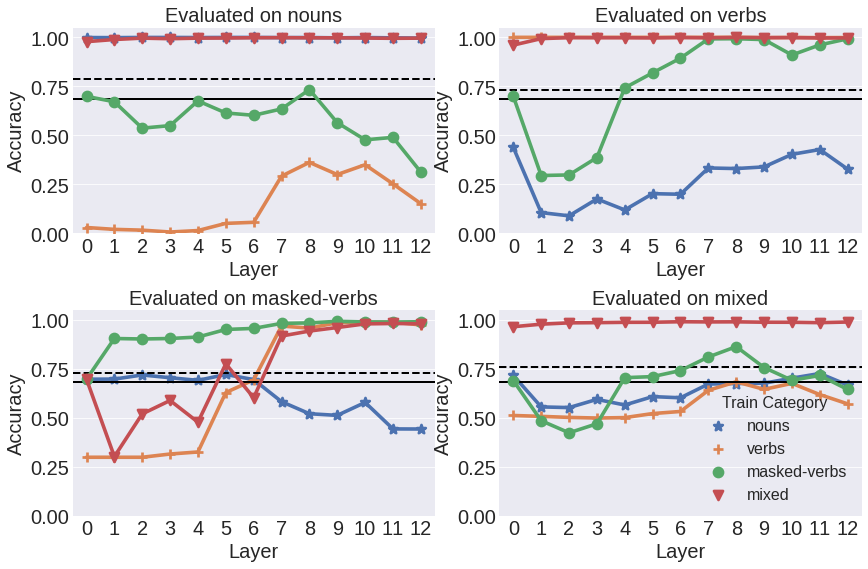}
        \caption{Probes cross-evaluation. Each plot corresponds to a test category, and colors correspond to the category used for training. Solid lines represent the percentage of majority-class (plural vs singular) tokens; dashed lines represent the percentage of majority-class tokens per lemma, averaged across lemmas.\looseness=-1
        }
        \label{fig:cross-class}
\end{figure}

In addition to comparing the angles of our diagnostic probes trained on different categories, we performed cross-evaluation of our trained diagnostic probes. In this setting, we trained probes on one category and tested them on the others. \cref{fig:cross-class} presents our cross-evaluation results.
The performance of probes evaluated in one category, but trained on another, again suggests that BERT encodes number differently across lexical categories. 
Interestingly, in the lower layer, the probe tested on nouns (top-left) guesses the wrong number systematically when trained on verbs, and vice-versa (top-right). 
This can be due to token ambiguity, as some singular nouns (e.g. ``hit") are also plural verbs. This is further evidence that the encoding might be different for nouns and verbs, though this analysis still cannot tell us whether this is true from our usage-based perspective. 
Additionally, the mixed results (\cref{fig:cross-class}; bottom-right), show it is possible to linearly separate both nouns and verbs with a single linear classifier trained on both categories, reaching perfect performance on all other categories, including masked-verbs (bottom-left).

\section{$\calV$-information and mutual information}
\label{app:v-information}
While a probing classifier's performance is often measured with accuracy metrics, in their analysis, \citet{pimentel-etal-2020-information} defined probing as extracting a mutual information. Formally, we write
\begin{equation}
    \MI(R; N) = \ent(N) - \ent(N \mid R)
\end{equation}
where $R$ and $N$ are, respectively, a representation-valued and a number-valued random variables.
The mutual information, however, is a mostly theoretical value---hard to approximate in practice.

To compute this, we must first define a variational family $\calV$ of interest; which we define as the set of linear transformations representable by \cref{eq:linear_pred}.
We can then compute the \vinformation{} as:\looseness=-1
\begin{equation}
    \vMI(R \rightarrow N) = \vent(N) - \vent(N \mid R)
\end{equation}
where $\calV$-entropies are defined as:
\begin{align}
    \vent(N) &= \inf_{\btheta \in \calV} \E_{n \sim p(n)} \log \frac{1}{\ptheta(n)} \\
    \vent(N \mid R) &= \inf_{\btheta \in \calV} \E\limits_{n, \br \sim p(n, \br)} \log \frac{1}{\ptheta(n \mid \br)}
\end{align}
This \vinformation{} can vary in the range $[0; \vent(N)]$; thus a more interpretable value is the \defn{$\calV$-uncertainty}, which we define here as:
\begin{equation}
    \vUnc(R \rightarrow N) = \frac{\vMI(R \rightarrow N)}{\vent(N)} 
\end{equation}

We note that the \vinformation{} lower-bounds the mutual information: $\vMI(R \rightarrow N) \le \MI(R; N)$.
It follows that, if we can extract some \vinformation{} from a set of representations, they contain at least the same amount of information in \citeposs{shannon1948mathematical} more classic sense.

\section{Attention Intervention} \label{sec:attention_intervention}

Formally, let $\bA^{l,h} \in \R^{T \times T}$ be a model's attention weights for a given layer $1 \leq l \leq 12$, a head $1 \leq h \leq 12$, and a sentence with length $T$.\footnote{Our analyzed model, BERT base, has 12 layers, and 12 attention heads in each layer.}
Further, we define a binary mask matrix $\bM^{l} \in \{0,1\}^{T \times T}$. 
We can now perform an intervention by masking the attention weights of all heads in a layer. Given a layer $l$:
\begin{equation}
   \widehat{\bA}^{l,h} = \bA^{l,h} \circ \bM^{l},  \quad 1 \leq h \leq 12 
\end{equation}
where $\circ$ represents an elementwise product between two matrices. 
Now assume a given sentence with cue position $p_{c}$, and with target position $p_{t}$.
In our intervention (i), matrix $\bM^{l}$ is set to all 1's except for $\bM^{l}_{p_{t},p_{c}} = 0$; the target's attention to the cue is thus set to 0. 
In intervention (ii), we set $\bM^{l}_{:,p_{c}} = 0$ and other positions to 1, which removes all attention to the cue.

\section{Removing random directions from representations}
\label{app:amnesic_baseline}
Removing directions from intermediate spaces could harm the model's normal functioning independently from removing our targeted property. We thus run a control experiment proposed by \citet{elazar2021amnesic}, removing random directions at each layer (as opposed to the specific directions found by our amnesic probes). This experiment allows us to verify that the observed information loss and decrease in performance do not only result from removing too many directions. To do so, we remove an equal number of random directions at each layer. The results are displayed in \cref{tab:amnesic-baselines} and show that removing randomly chosen directions has little to no effect compared to our targeted causal interventions.

\begin{table*}[h]
\centering
\resizebox{\textwidth}{!}{%
\begin{tabular}{lrrrrrrrrrrrrr}
    \toprule
     Layer & 0 & 1 & 2 & 3 & 4 & 5 & 6 & 7 & 8 & 9 & 10 & 11 & 12 \\
    \midrule
    Masked Verbs \\ \cmidrule(lr){0-0}
~~~~Number of Directions & 1 & 13 & 15 & 26 & 30 & 17 & 21 & 44 & 24 & 22 & 22 & 26 & 33 \\
~~~~Loss in Layers & 0.0 & 0.33 & 0.3 & 0.34 & 0.34 & 0.34 & 0.37 & 0.38 & 0.42 & 0.39 & 0.41 & 0.41 & 0.41 \\
~~~~Loss in Layers (Random) &  0.08 & 0.0 & 0.0 & 0.0 & 0.0 & 0.0 & 0.0 & 0.0 & 0.0 & 0.0 & 0.0 & 0.0 & 0.0 \\
~~~~NA Performance Drop & 0.04 & 0.01 & 0.01 & 0.01 & 0.01 & 0.0 & 0.01 & 0.0 & 0.0 & 0.09 & 0.29 & 0.33 & 0.23 \\
~~~~NA Performance Drop (Random) & 0.03 & 0.0 & 0.0 & 0.0 & 0.0 & 0.0 & 0.0 & 0.0 & 0.0 & 0.0 & 0.0 & 0.01 & 0.01 \\
    \midrule
    Nouns \\ \cmidrule(lr){0-0}
~~~~Number of Directions & 17 & 51 & 33 & 70 & 22 & 37 & 48 & 52 & 64 & 39 & 22 & 39 & 26 \\
~~~~Loss in Layers &  0.49 & 0.37 & 0.39 & 0.38 & 0.37 & 0.38 & 0.43 & 0.4 & 0.43 & 0.4 & 0.37 & 0.41 & 0.4 \\
~~~~Loss in Layers (Random) & 0.0 & 0.0 & 0.0 & 0.02 & 0.01 & 0.0 & 0.0 & 0.0 & 0.0 & 0.0 & 0.0 & 0.01 & 0.0 \\
~~~~NA Performance Drop & 0.32 & 0.32 & 0.27 & 0.29 & 0.28 & 0.29 & 0.22 & 0.09 & 0.04 & 0.0 & 0.0 & 0.0 & 0.0 \\
~~~~NA Performance Drop (Random) & 0.06 & 0.0 & 0.0 & 0.0 & 0.0 & 0.0 & 0.0 & 0.0 & 0.0 & 0.0 & 0.0 & 0.0 & 0.0 \\
     \bottomrule
\end{tabular}
}
\caption{Causal intervention results using both the default or random directions. For each category, we display the number of directions removed in each layer, the information loss resulting from amnesic interventions in each layer and the effect on the NA task. We also display the loss in layers and performance decrease on NA resulting from the removal of random directions as a control experiment.}
\label{tab:amnesic-baselines}
\end{table*}

\begin{figure*}
    \begin{subfigure}{\textwidth}
        \centering
         \begin{subfigure}[b]{0.28\textwidth}
             \centering
             \includegraphics[width=\textwidth]{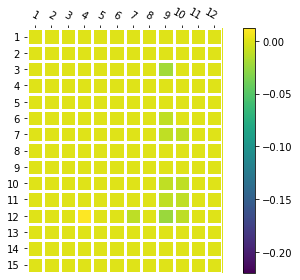}
         \end{subfigure}%
         ~
         \begin{subfigure}[b]{0.28\textwidth}
             \centering
             \includegraphics[width=\textwidth]{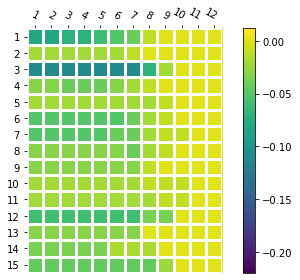}
         \end{subfigure}%
         ~
         \begin{subfigure}[b]{0.28\textwidth}
             \centering
             \includegraphics[width=\textwidth]{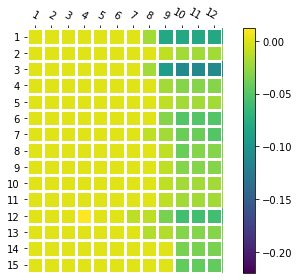}
         \end{subfigure}
         ~
          \begin{subfigure}[b]{0.11\textwidth}
             \centering
             \includegraphics[width=\textwidth]{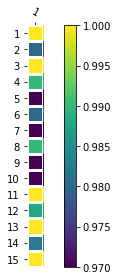}
         \end{subfigure}
        \caption{Cutting attention from the target to the cue only}
        \label{fig:attention-intervention-by-dist-matrices}
    \end{subfigure}
    
    \vfill{\textcolor{white}{space}}
    
    \begin{subfigure}{\textwidth}
        \centering
         \begin{subfigure}[b]{0.28\textwidth}
             \centering
             \includegraphics[width=\textwidth]{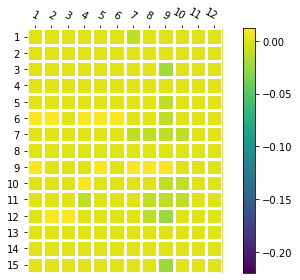}
         \end{subfigure}
         ~
         \begin{subfigure}[b]{0.28\textwidth}
             \centering
             \includegraphics[width=\textwidth]{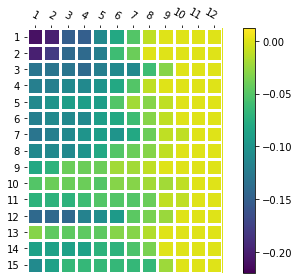}
         \end{subfigure}
         ~
         \begin{subfigure}[b]{0.28\textwidth}
             \centering
             \includegraphics[width=\textwidth]{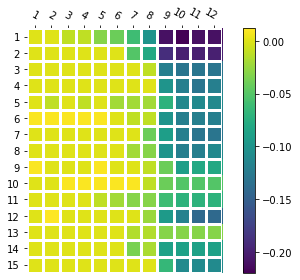}
         \end{subfigure}
         ~
          \begin{subfigure}[b]{0.11\textwidth}
             \centering
             \includegraphics[width=\textwidth]{imgs/base_scores_by_dist.png}
         \end{subfigure}
        \caption{Cutting attention from all tokens to the cue}
        \label{fig:attention-intervention-by-dist-matrices-all-cue}
    \end{subfigure}
    \caption{Agreement task performance drops resulting from attention interventions, as a function of linear distance between the cue and the target. The rows represent distances (from 1 to 15) and columns represent the intervened layers. Three conditions are tested: cutting attention only at current layer (left), cutting attention starting from current layer up to the last one (middle) and from the first layer to current layer (right). The color map on the far right represent agreement scores without intervention for each linear distance.}
\end{figure*}

\section{The effect of linear distance} \label{app:linear_study}

Here, we test whether the linear distance between the cue and the target influences the effect of attention removal. \cref{fig:attention-intervention-by-dist-matrices} shows that cutting attention from one layer has negligible effect over performance regardless of distance, which is in line with results from the diagonals of \cref{fig:attention-intervention-range-matrices}. When cutting attention from several subsequent layers (\cref{fig:attention-intervention-by-dist-matrices-all-cue}), we observe that performance drop depends on the linear position, and decreases when the model is not faced with short-range agreement. This is not surprising as many of the attention maps attend to surrounding tokens \citep{kovaleva-etal-2019-revealing}. Extensive analysis targeting individual attention heads (instead of cutting all attention from a given layer) is necessary to examine both their contribution to the model's successes, and their dependence on linear distance.

\end{document}